\documentclass[letterpaper, 10 pt, conference]{ieeeconf}  

\IEEEoverridecommandlockouts                              

\usepackage{graphics} 
\usepackage{epsfig} 
\usepackage{tabularx}
\usepackage{array,multirow,graphicx}
\usepackage{bm}
\usepackage{hyperref}
\usepackage{cuted}
\usepackage{capt-of}

\usepackage[dvipsnames]{xcolor}

\renewcommand{\paragraph}[1]{\vspace{.1em}\noindent\textbf{#1}.}

\title{\LARGE \bf
An Integrated Design Pipeline for Tactile Sensing Robotic Manipulators
}

\author{Lara Zlokapa$^{1}$, Yiyue Luo$^{1}$, Jie Xu$^{1}$, Michael Foshey$^{1}$, Kui Wu$^{2}$, Pulkit Agrawal$^{1}$, and Wojciech Matusik$^{1}$\\
\url{http://robohands.csail.mit.edu/}
\thanks{$^{1}$Authors are with the Computer Science and Artificial Intelligence Laboratory (CSAIL), Massachusetts Institute of Technology, 32 Vassar St, Cambridge, MA 02139, USA {\tt\small laraz@mit.edu}} 
\thanks{$^{2}$Author is with Lightspeed \& Quantum Studios, Tencent America, 12777 Jefferson Blvd, Building E, Los Angeles, CA 90066, USA} 
}%

\begin{document}

\maketitle
\thispagestyle{empty}
\pagestyle{empty}

\begin{strip}\centering
\vspace{-22mm}
\includegraphics[width=\textwidth]{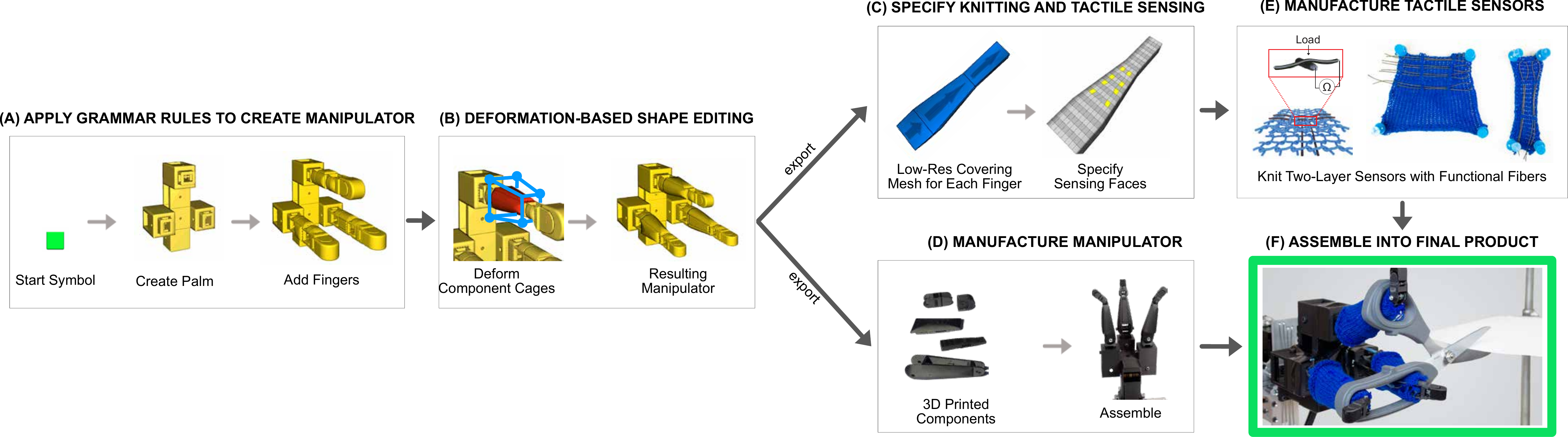}
\captionof{figure}{\textbf{Our Pipeline:} A manipulator is (A) \textit{generated using grammar rules} and (B) \textit{interactively re-shaped via cage-based deformation}. Next, the(C) user \textit{specifies touch sensor placement}. The sensors and hand components are (E) \textit{knitted} and (D) \textit{3D printed} prior to (F) \textit{manual assembly} of the manipulator.
\label{fig:workflow}}
\end{strip}

\begin{abstract}

Traditional robotic manipulator design methods require extensive, time-consuming, and manual trial and error to produce a viable design. During this process, engineers often spend their time redesigning or reshaping components as they discover better topologies for the robotic manipulator. Tactile sensors, while useful, often complicate the design due to their bulky form factor. We propose an integrated design pipeline to streamline the design and manufacturing of robotic manipulators with knitted, glove-like tactile sensors. The proposed pipeline allows a designer to assemble a collection of modular, open-source components by applying predefined graph grammar rules. The end result is an intuitive design paradigm that allows the creation of new virtual designs of manipulators in a matter of minutes. Our framework allows the designer to fine-tune the manipulator's shape through cage-based geometry deformation. Finally, the designer can select surfaces for adding tactile sensing. Once the manipulator design is finished, the program will automatically generate 3D printing and knitting files for manufacturing. We demonstrate the utility of this pipeline by creating four custom manipulators tested on real-world tasks: screwing in a wing screw, sorting water bottles, picking up an egg, and cutting paper with scissors.

\end{abstract}


\section{INTRODUCTION}

Currently designing robotic manipulators with tactile sensing is a time-consuming and manual process. One typically brainstorms designs to solve a specific task (or set of tasks), prototypes a selection of the brainstormed designs, chooses the most promising prototype(s), and repetitively iterates prototyping until a successful design is achieved. The majority of time is spent on design iterations driven by trial and error: several attempts may be necessary before a functional prototype is produced. These iterations often require both mechanical modifications (involving designing and testing new parts from scratch) and altering the topology of existing parts upon the realization that a part with different shape or size may be better suited for the task. Topology changes are time consuming because re-shaped pieces may no longer fit together physically, or component mates and parametric constraints in the CAD program may break after dimensions are modified. When this happens, a human must re-model components, identify and fix parametric constraints, and manually re-connect CAD assembly pieces to re-assemble the manipulator model.

Integrating tactile sensors in a robotic manipulator further complicates the already human-labor intensive design process. Many tactile sensors~\cite{yuan2017gelsight, BubbleGripper} are bulky and cannot be simply added on top of existing designs. Instead, manipulators must be designed around them. This adds further geometric constraints on component interface sizing to the already tedious process of modifying part geometries using traditional CAD programs. From start to finish, depending on task and design complexity, it may take months or years to produce a high-quality robotic manipulator.

We propose a pipeline with an interactive user interface to streamline the design and manufacturing process which is illustrated in Fig. \ref{fig:workflow}. Our pipeline enables the user to design task-specific, cable-driven robotic manipulators with pressure sensing in a matter of minutes. Using our open-source collection of modular sub-component 3D models and the proposed grammar rules for assembly, users can quickly create many different robotic manipulator CAD models. 
 Because the connections between sub-components are encoded in the grammar rules, a complete 3D model of the manipulator updates in real-time as the user applies grammar rules. No manual assembly of components in a CAD program is required. Then, using an intuitive, cage-based deformation method, the user may topologically deform (i.e., lengthen, widen, and otherwise distort) the manipulator models according to the desired finger and hand shape and size. If needed, at this stage, users can identify small regions for placing tactile sensors on the manipulator's surface. The sensors are built into a knitted cover that conforms around the manipulator like a glove. 

Our design pipeline \textit{guarantees} that each manipulator design can be manufactured. Once the manipulator is virtually completed, our program \textit{automatically} generates manufacturing files for 3D printing manipulator components and files for \textit{automatically} manufacturing the tactile glove via an industrial knitting machine. By removing many manual and time-consuming steps in the traditional approach (e.g., modeling CAD components, laboriously assembling components in CAD, adapting CAD models during geometry changes, and tediously integrating touch sensors), our proposed pipeline enables designers to focus on improving form rather than fixing functionality.


We evaluated the efficacy of our design pipeline on four tasks chosen to demonstrate the breadth of possible designs and the ability to integrate tactile sensors. We designed and manufactured four separate manipulators to (1) pick up an egg, (2) screw on a wing screw, (3) sort water bottles, and (4) cut paper with scissors.

\section{RELATED WORK}

\paragraph{Manipulator Design}
Existing robotic hands such as the Shadow Hand, DLR Hand~\cite{grebenstein2012hand}, UW Hand~\cite{xu2013uw}, RBO Hand 2~\cite{deimel2016novel}, and others~\cite{rus2015design,piazza2019century} have been designed using traditional methods for a specific set of tasks. Changing the end task would require complete redesign which is likely to consume months if not years. For instance, large joints may not be manufacturable at a small scale, or task-specialized manipulator fingers may not be cross-applicable for other tasks, requiring new brainstorming, testing, and specialized design.

Modular robot design is an effective way to generate various robot structures from a small library of base components. It has been applied to generate whole robots (\cite{jing2018accomplishing, bi2001concurrent, chen1995determining}) and modular hands, including ModGrasp~\cite{modgrasp}, OpenMRH~\cite{openmrh}, and NSU's sensorized, pneumatic robotic hand~\cite{NSU}. These hands, though modular, generally rely on a single standard finger model that tessellates to extend or shorten the finger. Most similar to ours is the Yale OpenHand that offers a library of components for assembly~\cite{openhand_mod, openhand_opt}. However, previous works have a limited component selection, resulting in a limited set of possible topologies. Additionally, they only explore the discrete topology space of robot designs. In contrast, our system considers both the discrete topology of the manipulator and the continuous geometry of each component thus providing a richer design space. Finally, none of these modular hands employ grammar or allow for custom deformation.

Grammar-based design paradigm has previously been employted to generate simulated multi-pedal robots~\cite{zhao2020robogrammar}, mathematically model the self-assembly of robotic systems~\cite{klavins2004graph}, create IKEA cabinets and tables~\cite{lau2011converting}, and generate passive dynamic brachiating robots~\cite{brachiating_robots}. However, grammar driven design  not been employed for creating manipulators nor has it been tested in the real-world. To the best of our knowledge, our work consists of the first demonstration of an integrated computational framework for the design of robotic manipulators and sensor placement.

\paragraph{Tactile Sensing}
Many commonly available tactile sensors have form-factor and compliance restrictions that impact the geometry and structure of manipulator design. The Robonaut 2 Hand has rigid sensors built into the palmar side of the hand's phalanges~\cite{RobonautSensors,ihrke2010phalange}. The iCub Hand is conformally covered with flexible PCB acting like a capacitive pressure sensor at the finger tips~\cite{iCubSensors}. The RBO 2 Hand is wrapped in liquid metal strain sensors to calculate the deformation and extrapolate contact with the grasped objects~\cite{RBO2Sensors}. 
Finally, some hands may be fitted with BioTac sensors~\cite{BioTac}, ready-made sensorized fingertips that are fitted on the robotic hand in place of the original hand's finger tips. These sensors are available in a single size and their form factor cannot be altered if a different manipulator topology is required. Other common tactile sensors are the Tekscan Grip system~\cite{tekscan_2018}, Gelsight and other vision-based silicone sensors~\cite{yuan2017gelsight,Li2014,Yamaguchi2016, Tsinghua_sensor}, and the biomimetic multimodal sensor~\cite{Wettels2011,Parke1500661}. However, none of these can both be designed and manufactured in a computer-automated manner and conformally cover the robot hand of complex geometry. Electronic skins~\cite{Chortos2016PursuingPE, boutry2018hierarchically}, while more flexible and adaptable, have not been scaled up to larger sizes due to the delicate manual manufacturing processes. 


In this work, we incorporate computational design and digital fabrication of knitted pressure sensing matrices to conformally cover our manipulators in a scalable, cost-efficient manner. We hope this will enable broader exploration of tactile sensing for object manipulation.

\section{Design Workflow}
\label{sec:design-workflow}

Our design workflow is summarized in Fig. \ref{fig:workflow}. 
We designed a library of components (Fig.~\ref{fig:components}) that can be combined using the proposed context-sensitive grammar (Sec.~\ref{ssec:grammar}) to create a diverse family of manipulators. 
From this discrete design space, a manipulator topology is chosen. Next, the component shapes can be refined (Sec. \ref{ssec:deformation}) to increase manipulator's suitability for the desired task. 
The design requirements of grammar drive composition, ease of shape deformation, and the guarantee on manufacturing impose constraints on component designs that are discussed in Sec.~\ref{ssec:component_design}. 
Finally, Sec.~\ref{ssec:sensing} details how the user can specify the location of touch sensors. 

\subsection{Context-Sensitive Grammar}
\label{ssec:grammar}

\begin{figure}[t]
    \vspace{0.4em}
    \centering
    \includegraphics[width = \columnwidth]{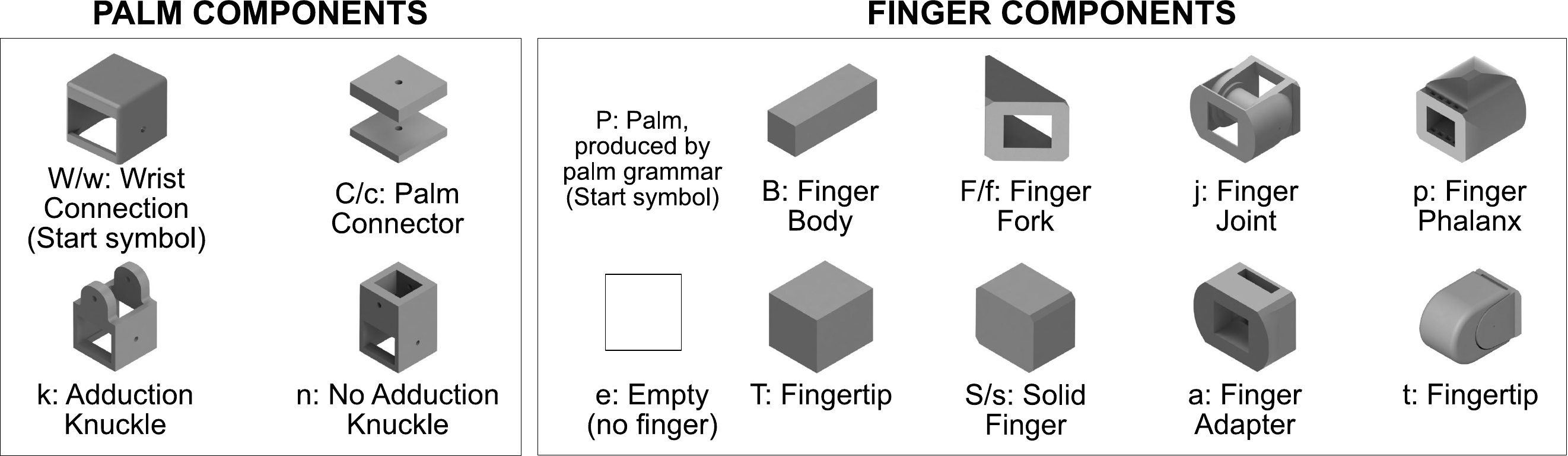}
    \caption{\textbf{3D models of the grammar's components with associated symbols.} Capital letters indicate that the component is a non-terminal symbol, while lowercase letters indicate a terminal symbol.}
    \label{fig:components}
\end{figure}

\begin{figure}[t]
    \centering
    \includegraphics[width = \columnwidth]{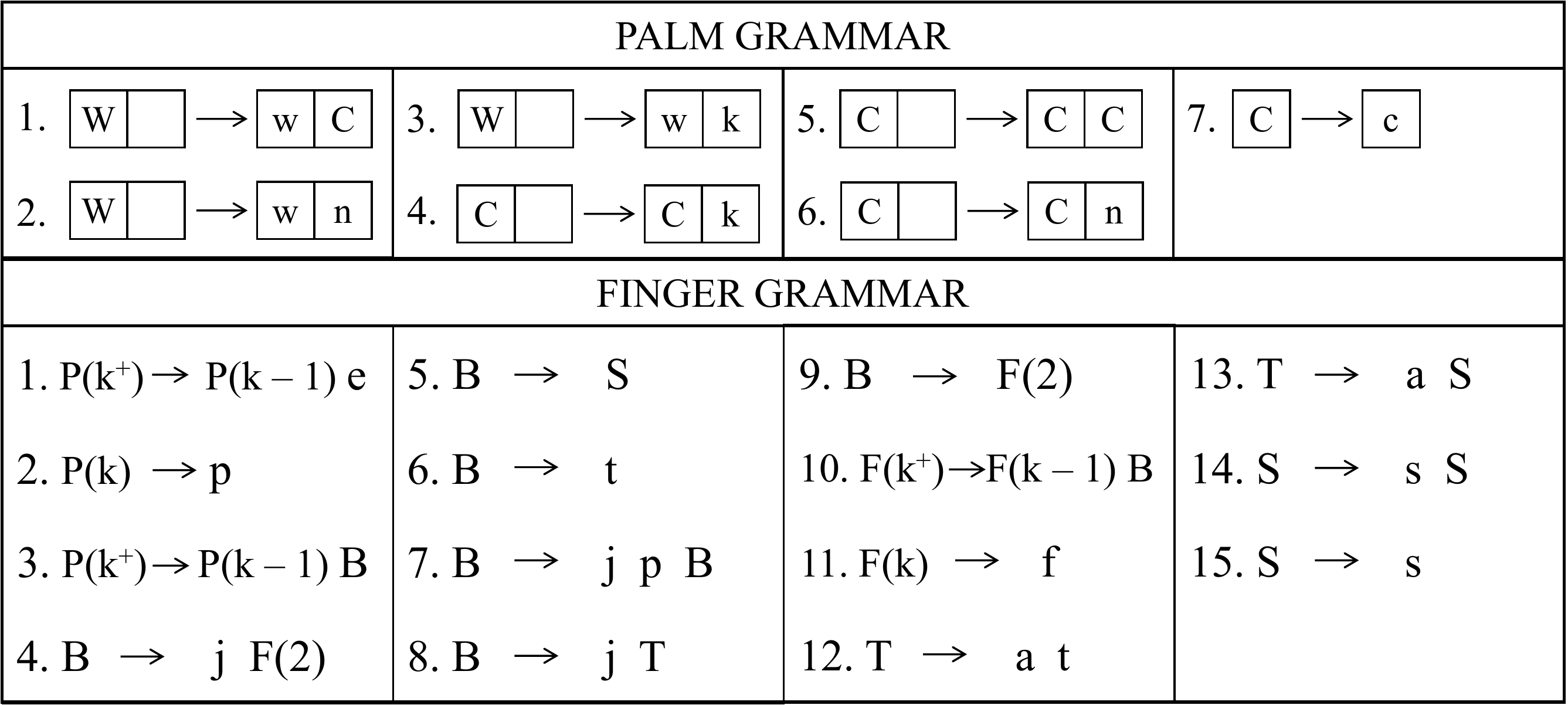}
    \caption{Grammar expansion rules for constructing fingers and palms. The palm grammar is defined on a grid layout and the finger grammar is a parametric grammar where the palm node ``P'' and fork node ``F'' contain an integer parameter $k$ to denote the number of rule expansions can be made on the node. $k^+$ means that rule can be applied only when $k$ is positive.}
    \label{fig:grammar}
\end{figure}

We represent a manipulator assembly design as a graph where each node corresponds to a physical sub-component and each edge encodes a connection between two sub-components (e.g., relative rotation, translation, etc.). This choice of graph representation guarantees that each assembly has a unique graph, and each graph corresponds to a unique assembly. The task of generating diverse manipulator designs therefore reduces to generating diverse graphs. 

\begin{figure}[t]
    \centering
    \includegraphics[width = \columnwidth]{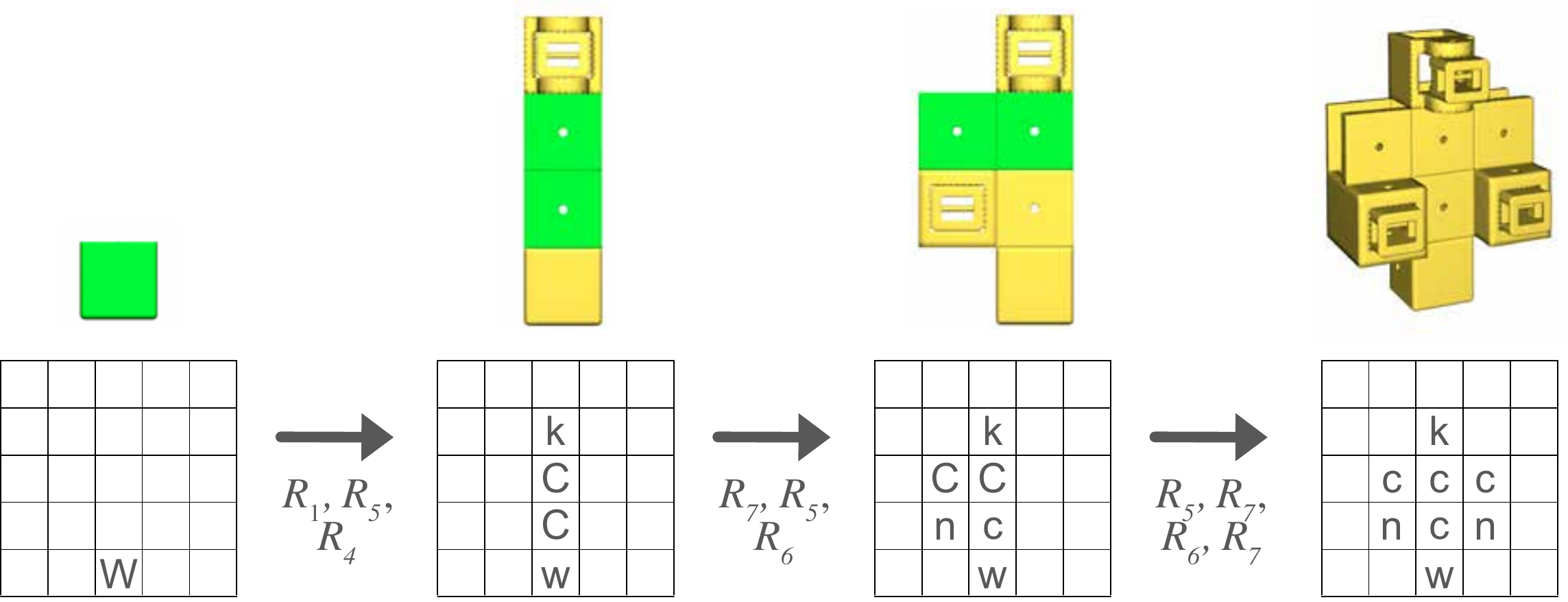}
    \caption{The palm grammar rules are applied to grow the start symbol (W), add connector components (C), and attach knuckles (k and n) to create the grid-based water bottle palm. Green components are non-terminal, and yellow components are terminal. Rule numbers ($R_{\#}$) correspond to Fig. \ref{fig:grammar}.}
    \label{fig:palm_grid}
\end{figure}

\begin{figure}[t]
    \centering
    \includegraphics[width = \columnwidth]{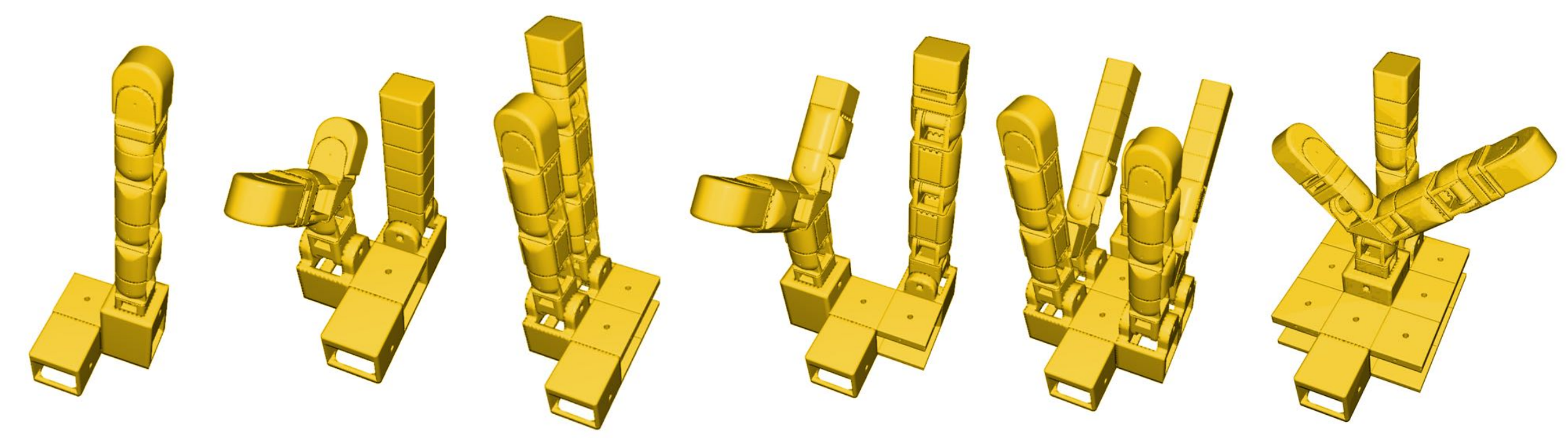}
    \caption{A diverse sample of manipulator designs generated by combining components using the proposed grammar rules. The designs shown in the figure are from the stage before deformation.}
    \label{fig:possibilities}
\end{figure}


Our manipulator grammar consists of two sub-grammars, a \textit{palm grammar} and a \textit{finger grammar}, with rules defined in Fig. \ref{fig:grammar}. The \textit{palm grammar} generates palms of different sizes, shapes, and numbers of finger slots. Once the palm grammar produces a palm, the user proceeds with the \textit{finger grammar} to grow fingers from the palm. Each grammar consists of:
\begin{enumerate}
  \item \textit{Terminal symbols} (noted as lowercase letters). These represent the nodes and edges of a graph.
  \item \textit{Non-terminal symbols} (noted as uppercase letters). These represent sub-assemblies or sections of a graph.  
  \item \textit{A start symbol}. A non-terminal symbol that initializes the design.
  \item \textit{Expansion rules}. These convert non-terminal symbols into other non-terminal and terminal symbols. They allow the creation of many different graphs based the order and selection of rules applied.
\end{enumerate}
The terminal and non-terminal components used to create the manipulators are shown in Fig.~\ref{fig:components} with their associated letter symbols. Note that we only show the components required to make the manipulators in this paper; the ``library" of components can be augmented as desired.

\noindent \textbf{Palm Grammar}: To design diverse palms, we compose components shown on the left side of Fig. \ref{fig:components}. These components can be connected in a planar grid using the palm grammar rules to generate palms of varying shapes (see Fig. \ref{fig:palm_grid}). It should be noted that each palm grammar rule can be applied in three configurations rotated by $90l (l\in{1, 2, 3})$ degrees to expand the palm in different directions. For example, a knuckle node (k) can be connected to either left, right, top and bottom to a connector node (C). Once the palm has been built, it serves as a start symbol for the finger grammar to attach fingers if desired.

\noindent \textbf{Finger Grammar}: Similar to the structure of human fingers, the \textit{finger grammar} combines finger components depicted in Fig. \ref{fig:components} linearly: components are added distally to the fingers to ``grow" them in length until termination with a fingertip.

Fig. \ref{fig:possibilities} shows a few grammar-generated manipulator designs. 
With only thirteen components, our grammar can produce myriad manipulator configurations with different palm or finger shapes. On the order of $10^{8}$ unique fingers can be generated from fifteen finger expansion rules and six terminal finger components, assuming the fingers are constrained to lengths of at most three segments. Restricting the palm to a three by three grid (only for calculation purposes) with up to six fingers, at least $10^{49}$ unique hands exist within our constraints. Such a broad design space can be drastically increased with additional grammar pieces. To efficient exploration of the vast design space, we developed an interactive design interface (Fig. \ref{fig:workflow}(A)).

\subsection{Deformation-Based Geometry Shape Design}
\label{ssec:deformation}
Once the hand's topology is established, our pipeline proceeds to the shape refinement phase, where the user may change the geometry of indivdual components. While the grammar generates discrete manipulator topologies, the specific dimensions of the manipulators may be sub-optimal for the desired task. For instance, it may be beneficial for the phalanges to be longer or for the fingers to taper. Our geometric deformation method supplements the discrete grammar-based designs by enabling users to quickly and intuitively vary the manipulator's shape in a continuous manner to further optimize their design. To allow users to easily make design deformations that span multiple shape dimensions, its logical to use a low-dimensional design parameterization.

In traditional CAD modeling, users would manually parametrize each feature's dimensions, a mistake prone process. Inspired by Xu et al. \cite{Xu-RSS-21}, we apply cage-based deformation to parameterize the manipulator.
This technique encloses each high-resolution subcomponent mesh into a \textit{cage}-like coarse and cuboid shape mesh. The shape of the enclosed subcomponent can be altered by simply moving the cage vertices to scale, shear, and taper while guaranteeing that connections with other surrounding subcomponents are preserved, thereby ensuring manufacturability.
We adapt Xu et al.'s method to build an intuitive user interface (UI) for interactively modifying the shape of the manipulator as shown in Fig. \ref{fig:workflow}(B). The UI allows users to manipulate the cage mesh vertices of each subcomponent, deforming the manipulator's underlying high-resolution mesh.

\subsection{Grammar Component Design}
\label{ssec:component_design}
Each grammar component is associated with three meshes. First, there is a high-resolution mesh used for 3D printing and manipulator renderings. Second, a coarse, cuboid \textit{cage} mesh encloses each component and is used to specify deformation. Third, to generate the knitted sensors, there is another low-resolution, cuboid mesh that approximates the associated high-resolution mesh shape. Both the high resolution and knitting meshes are deformed by changes to the \textit{cage} mesh. Appendix \ref{app:meshes} contains more details on these meshes.

\begin{figure}[t]
    \vspace{0.4em}
    \centering
    \includegraphics[width = \columnwidth]{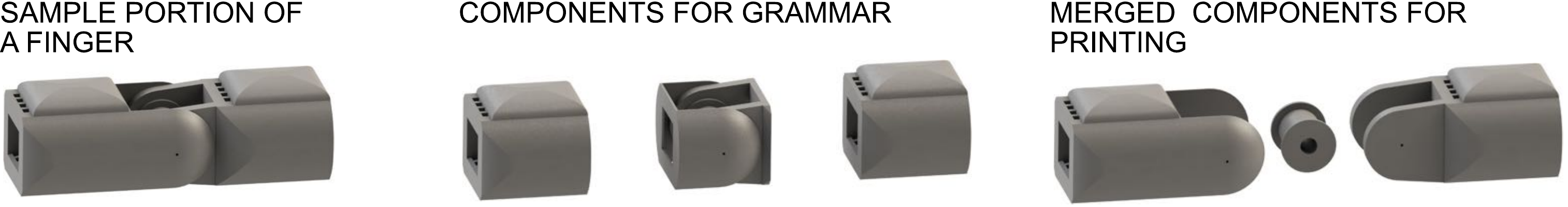}
    \caption{Grammar components correspond to mechanical systems rather than physical parts. They can be combined into their proper physical parts that are suitable for 3D printing after deformation.}
    \label{fig:combining}
\end{figure}

To preserve the mechanical relationships between parts during deformation, components correspond to mechanical systems (e.g., static phalanges or dynamic joints) rather than to physical parts (e.g., phalanges with joints) as shown in Fig. \ref{fig:combining}.
In the figure, two phalanges that are attached by a joint are separated into three grammar components (Fig. \ref{fig:combining}, middle): two phalanx shafts and one pin joint. The joint consists of three physical pieces: the distal end of one phalanx, a pin, and the proximal end of another phalanx. The joint component can only be scaled uniformly or axially during cage-based deformation to ensure it functions as a joint. Shearing or scaling in any other axis will result in elliptical pins, which cannot pivot and therefore do not work, breaking the mechanical relationship between the components in the joint. In contrast, the shaft (phalanx) of a finger may be stretched and sheared in almost any manner without impeding its function. Since deformation cages enclose and control each component, dividing functional aspects into separate components with different deformation considerations allows users to maximally deform the assembly without compromising functionality.
This choice of part division, combined with careful grammar rule selection, ensures that any design can be manufactured. 

\subsection{Tactile Sensing Cover Design}
\label{ssec:sensing}
Given the embedded cuboid mesh from the deformation stage, we offer a design interface based on the stitch meshes framework~\cite{Yuksel2012:stitchmeshes, narayanan2019visual} for users to place sensors and generate corresponding knitting files (Fig. \ref{fig:workflow}(C)). To ensure the manipulator can wear the knitted cover, users specify the sides of the finger that the sensor wraps around. From this, our system generates a planar knitting pattern with a quad dominant stitch mesh based on the pattern's edge length. Each stitch face in this quad-dominant stitch mesh \cite{Yuksel2012:stitchmeshes} represents a real stitch in the knitted cover: a quad represents a knit and a pentagon represent an increase or decrease stitch to make the knitted structure conform to the component mesh. Users select a stitch face to place a sensor. Finally, our system automatically generates machine knitting files: it traces a knitting path and specifies stitches based on if the face is a knit, increase/decrease, or a sensor.

Our knitted sensors were manufactured based on \cite{luo2021learning}. Each knitted cover requires two layers, one with horizontally and the other with vertically integrated piezoresistive fibers. By overlapping the two layers, we form a sensing matrix: sensing points are located at the intersections of the orthogonally overlapped fibers where the piezoresistive nanocomposite is sandwiched by two conductive electrodes. The tactile sensors convert pressure stimuli into electric signals, which a customized read-out circuit acquires.


\section{FABRICATION}
The grammar subcomponents were designed in SolidWorks then imported into the proposed program. As shown in the accompanying video, in the program, we manually applied grammar rules to create manipulator graphs, deformed the manipulators to the desired shape, and specified all tactile sensing points. The program generated STL and DAT files for 3D printing the hand and knitting the sensors. After manufacturing, the printed pieces were assembled, cables strung, and sensors stitched closed over the manipulators. Each manipulator was mounted on a motor box of Dynamixel motors, which was then mounted on a UR5 arm.

\subsection{Hand Structure}
The program-generated STL files were 3D printed on a Markforged printer using Onyx, a micro-carbon fiber filled nylon, and assembled after printing. Bushings were added in the palm using holes in the Connector subcomponents to guide the cables. Spectra cables were threaded from each motor, through the manipulator, to the finger joints, and back to the motors that they originated from,, creating a fully-actuated, closed-loop cable drive. After threading and tying all the cables, the system was tensioned by sliding each motor down a short track on its mount until the cables on the motor became sufficiently tight (approx. 20-30 N). The motor was then tightened to prevent it from sliding.

\subsection{Knitted Sensor}
Given the generated knitting instructions, we used a digital knitting machine (SWG091N2, Shima Seiki) to knit the tactile sensing cover by integrating coaxial piezoresistive fibers into the textile. A customized read-out circuit interprets the electrical signals caused by pressure to the sensor.

\section{APPLICATION AND RESULTS}

\begin{figure}[t!]
    \vspace{0.4em}
    \centering
    \includegraphics[width = \columnwidth]{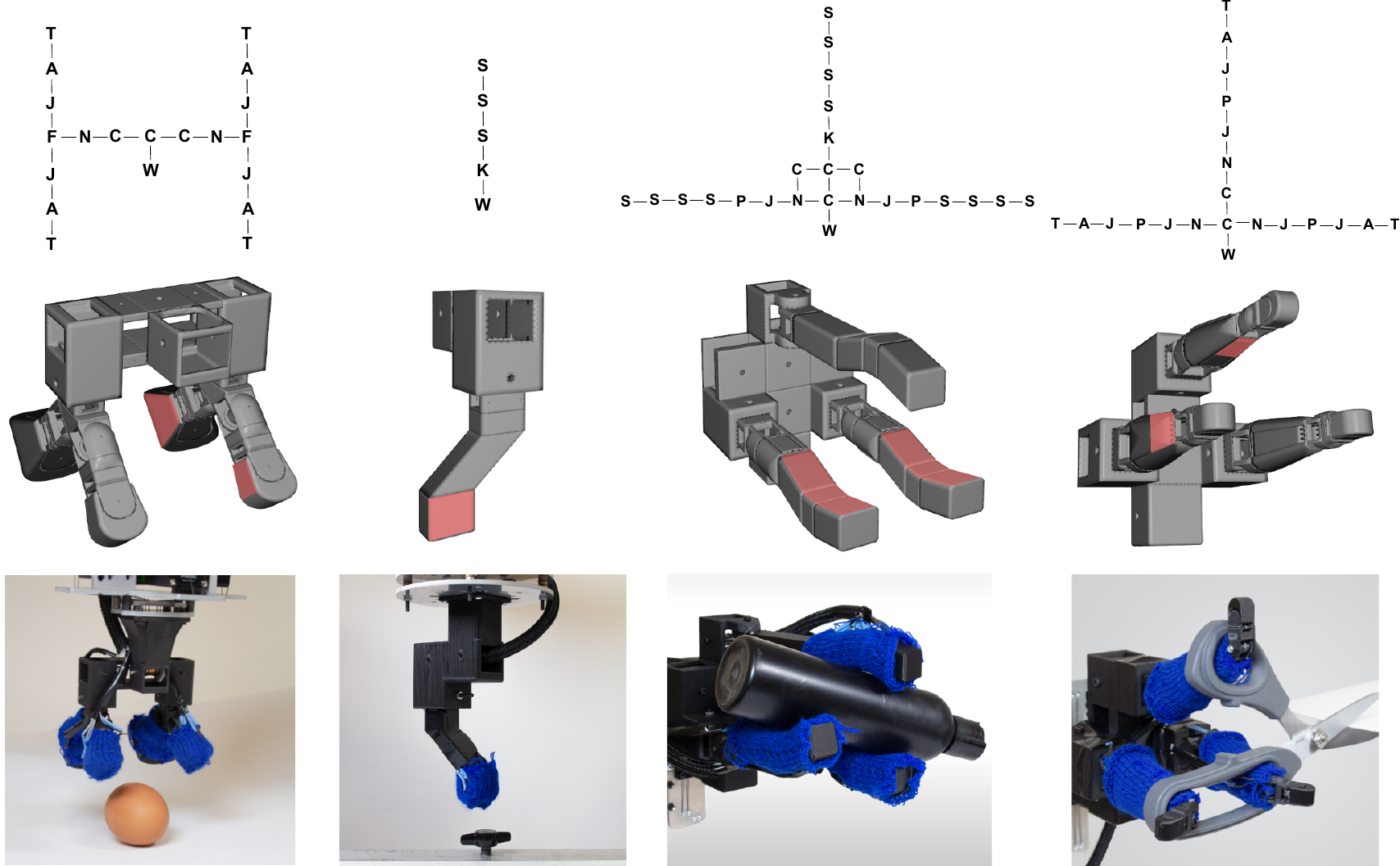}
    \caption{\textbf{Graph, 3D model, and image of each manipulator.} Top row: Grammar rules were applied to create graphs. Middle row: The graphs were modeled and topologically deformed via cage deformation to produce manipulators; user-specified sensor locations are shown shaded in red. Bottom row: Images of the fabricated and tested manipulators.}
    \label{fig:tasks}
\end{figure}

A separate manipulator was manufactured for each of the four tasks (detailed below), mounted on a UR5 arm, and controlled with a simple ruled-based policy (Appendix \ref{app:fsm}) to test the success of the design pipeline. To ensure the control policy works despite the sensor noise, we processed touch readings as follows: (i) At the start of each task, the sensor readings were normalized to be zero mean by computing the average sensor reading from the first fifteen time steps when the manipulator was not in contact with any object. (ii) Sometimes, due to shifts in the fabric, readings may measure negative pressure; we clipped the readings at 0. (iii) The maximum sensor value detected on the surface of each finger at any given time was used to guide controls: these readings correspond to the firmest points of contact with the sensors. Finally, due to shifts between the two knitted sensor layers during handling and storage, the contact between layers changed between testing sessions. Therefore, all pressure threshold values used during manipulator control were experimentally determined and tuned before each recording session. Fig. \ref{fig:tasks} shows the graph, 3D model, and manufactured manipulator for each task. The associated video contains task demonstrations, and Appendix \ref{app:sensor_graphs} depicts sensor readings.

\subsection{Picking up an Egg}
Eggs require delicate handling. Picking an egg tests the sensitivity of the sensors and the pipeline's ability to generate a manipulator that can grasp an object securely and carefully.
\subsubsection{Task description}
A sensorized four-finger hand picks up an egg from a table, shakes it to demonstrate the secure grasp, and places it back on the table.
\subsubsection{Design}
This design was selected based on engineering intuition as the most secure method of holding an egg. The manipulator fingers are mounted on a forked finger with angled joints to conform around the egg when they bend. The two lower fingers take advantage of our continuous deformation: they are are significantly wider at the tip to better cradle the wider base of the egg. The deformed width of the manipulators was determined by measuring the egg, and the design required only one attempt. Six sensors are located on the inner surface of each of the four fingers that contacts the egg.
\subsubsection{Control}
From a set location, the handed closes its fingers around the egg until the sensors on most of the fingers reach an experimentally-determined pressure threshold. It follows a hard-coded sequence to shake and release the egg.
\subsubsection{Performance}
\begin{figure}
    \vspace{0.4em}
    \centering
    \includegraphics[width = \columnwidth]{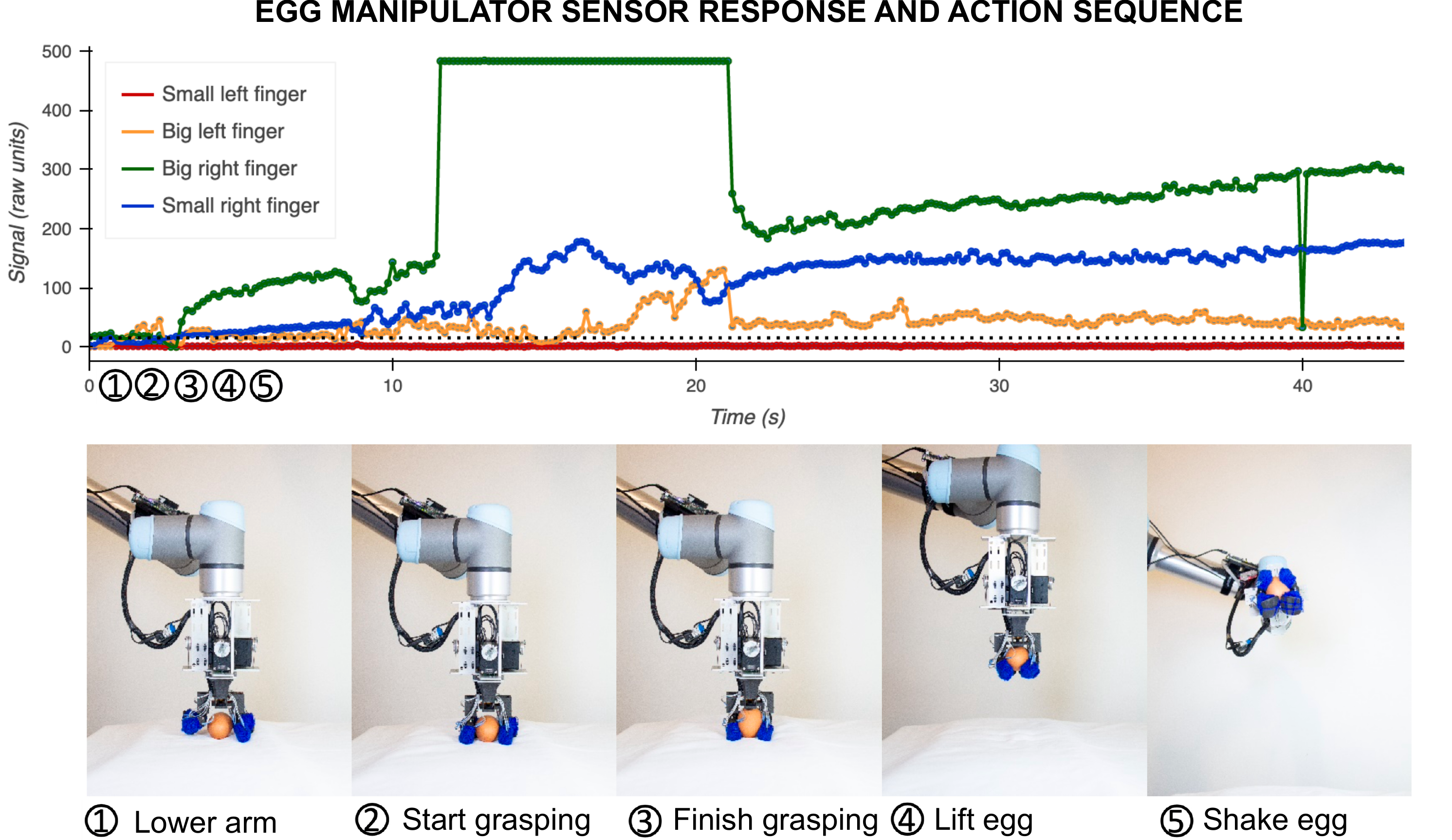}
    \caption{A typical egg picking action sequence of the manipulator alongside the maximum readings from the tactile sensors on each finger. We register the grasp as successful if three of the four readings exceed the threshold.} 
    \label{fig:eggpicker_tactile}
\end{figure}

The manipulator picks up the egg and grasps it securely so that the egg does not move when the robotic arm shakes it. It releases the egg without any breakage. We note that grasp is reliable enough that we never broke or dropped any eggs during testing. A sample grasping sequence with a tactile sensing signal is shown in Fig.~\ref{fig:eggpicker_tactile}.

\subsection{Screwing on a Wing Screw}
\subsubsection{Task description}
In our demo, a single, sensorized, rigid finger mounted on a UR5 arm screws a wing screw into a hole until tight. Requiring no moving parts (other than the UR5 wrist), this is one of the simplest designs that the design space can produce.
\subsubsection{Design}
Consisting only of a single, rigid finger, the manipulator is designed to minimize the number of moving parts (i.e., only rotation of the UR5 wrist) while ensuring that it is possible to screw in the wing screw. The base of the finger is lengthened and offset (sheared, using our cage deformation) to prevent the finger from colliding with the nut while the fingertip contacts the wing. The width of the finger accommodates several wing screw sizes. Sensors are located on the contact surface between the wings of the wing screw and the finger.

This design required only two iterations. The first iteration was a straight finger, which required more complicated control of the UR5 arm. The second iteration (seen in Fig. \ref{fig:components}) offset the fingertip from the axis of the wing screw so that the only required motion is rotation of the UR5 arm wrist.
\subsubsection{Control}
From a set starting point, the sensorized finger maintains contact with the wing screw so that when the UR5 wrist rotates, the wing screw is tightened. Because the wrist of the UR5 arm cannot rotate more than one full rotation, the robot is programmed to (1) perform half a clockwise rotation to twist the bolt, (2) lift the manipulator up, (3) rotates back 180$^{\circ}$ to reset the wrist angle, then (4) lower the manipulator to repeat the screwing in process until the sensor reads that a force greater than a pre-determined threshold is required to rotate the wing screw. This indicates that the wing screw is screwed in.
\subsubsection{Performance} Upon tuning of the pressure threshold, the manipulator successfully performs the task.

\subsection{Sorting Bottles}
\subsubsection{Task description}
A manipulator ``weighs" a water bottle to detect if it is empty or full. If full, it asks that the cap be unscrewed; then it pours water from the bottle. Empty bottles are discarded in a bin.
\subsubsection{Design}
Cage deformation allows the manipulator's fingers to be curved to better grasp round bottles by lengthening and shearing each of the solid (s) components. Before creation, the design concept was tested with three pencils (representing the three fingers) to determine if the bottle could be tilted using only three degrees of freedom. Once verified, the design was created using the proposed framework in a single attempt.
\subsubsection{Control}
The three-finger manipulator ``weighs" a water bottle by balancing it on two fingers and comparing the pressure detected to a pre-determined threshold. To pour water from the bottle, the two bottom fingers flex and extend, tilting the bottle while the top finger abducts to balance the bottle. To release, the UR5 arm moves the manipulator over the discard bin, and the fingers extend by a pre-set amount.
\subsubsection{Performance}
The manipulator was moderately successful at detecting if the water bottles were full or empty, pouring water from the full ones once the cap was unscrewed. It successfully deposited empty bottles in the discard bin.

\subsection{Cutting Paper with Scissors}
This scissor manipulator demonstrates grip adaptability
(it dons most shapes of office scissors) and dexterity in  handling the scissors. Sensor feedback determines when each cut has been completed.
\subsubsection{Task description}
A three-finger manipulator first dons scissors then cuts with them. If a hard material is placed between the scissor blades instead of a sheet of paper, the tactile sensors detect that excessive force is required when attempting to cut the material, and it will stop cutting.
\subsubsection{Design}
Since scissors are created for human hands, a design with a thumb and two opposing fingers was selected. To accommodate finger holes of varying shapes and sizes in office scissors, the ``pointer" and ``middle" fingers abduct to ``expand" to fill the finger hole, stabilizing the scissors. The fingers were tapered using cage deformation to prevent the scissors from slipping towards the base of the palm. Flexed distal joints ensure that the scissors do not slip off the fingers.

This design required the most iterations to cut with any brand of office scissors. Three configurations were tested, including two types of non-articulated fingers for the larger scissor handle hole. The configuration seen in this paper required two iterations, where the fingertip was narrowed to accommodate narrower scissor handle holes.
\subsubsection{Control}
To don the scissors, the robot fully extends the fingers. The scissors are placed on the robot hand, then the robot abducts the two fingers in the larger scissor handle hole and flexes all distal joints for a secure grasp. To cut, the robot opens the scissors by spreading its thumb, moves the hand forward a prescribed amount, and flexes the thumb to close the scissors until the pressure threshold is reached.
\subsubsection{Performance}
The manipulator is able to don a variety of office scissors and cut paper with them, stopping cutting when the paper has been cut. It detects when a hard surface is put between the scissor blades and does not cut it. The accompanying video demonstrates the scissors cutting through paper and rejecting cutting hard acrylic sheets.

\section{Discussion and Conclusion}
In this paper, we presented a design pipeline for creating a variety of robotic manipulators and demonstrated application of the design method with four manipulators for four tasks. A user interface enabled the user to design the manipulator's morphology using a context-sensitive grammar, topologically deform it, and specify tactile sensing points. The program then automatically generates files for manufacturing, and the user assembles the manipulator with the knitted sensors.  This self-contained method simplifies the manipulator design process by providing a grammar that allows the user to flexibly arrange and re-arrange components in a speedy manner while ensuring that all component configurations result in manufacturable designs. Additionally, it shortens the redesign time by providing a basis of pre-tested components, giving the user confidence that initial designs will perform.

While the pipeline performed well for both design and manufacturing, we hope to improve the sensors and motor box. Though functional and easy to manufacture, the sensors require substantial manual tuning in every repetition of each of the four tasks. Improving the accuracy and reliability of sensors is an important area of future investigation. We also found that the manipulator cables were subject to breakage at the motor attachment point at high loads. In principle, this problem can be resolved by increasing the cable diameter and improving the attachment method so that the cable does not pass over any sharp $90^{\circ}$ bends at the motor.

Our design pipeline has applications beyond manual design. The computer-friendly graph and grammar representation enables interfacing with ML, AI, and other optimization and simulation software. For instance, it may be integrated with the already-developed AI-driven geometry and control optimization ~\cite{Xu-RSS-21} to quantitatively optimize topology and control for user-specified manipulator configurations. Alternately, it may be integrated with an algorithm that methodically searches and simulates the design space to determine the optimal manipulator. The ease with which this program can be integrated with other computational processes opens many opportunities for co-optimization on the control and simulation fronts. In the future, with this pipeline, it may be possible for a program to automatically create optimized robotic manipulators with automated manufacturing and computer-generated controls algorithms in a matter of hours without any human involvement.


\section*{ACKNOWLEDGMENT}
This work was supported by Toyota Research Institute, Defense Advanced Research Projects Agency (FA8750-20-C-0075) and an Amazon Robotics Research Award. 

\bibliographystyle{IEEEtran}
\bibliography{IEEEabrv,bibliography}

\clearpage
\begin{appendices}

\section{Manipulator Generation}
\label{app:graphs}
In this section, the grammar rules used for each manipulator are referred to by number only where $R{p}$ refers to palm grammar rules and $R{f}$ refers to finger grammar rules. The rule associated with each rule number can be found listed in Fig. \ref{fig:grammar}. In order of operation, the following grammar rules were applied for each manipulator:

\subsection{Egg manipulator.}
\noindent Palm: $R_{p}1, R_{p}5, R_{p}4, R_{p}7, R_{p}5, R_{p}4, R_{p}7, R_{p}7.$\\
For each finger: $R_{f}3, R_{f}9, R_{f}10, R_{f}8, R_{f}12, R_{f}10, R_{f}8, \allowbreak R_{f}12, R_{f}11$.\\
After developing all fingers: $R_{f}2$.

\subsection{Wing screw manipulator.}
\noindent Palm: $R{p}2$.\\
Finger: $R_{f}3, R_{f}5, R_{f}14, R_{f}14, R_{f}15, R_{f}2.$

\subsection{Water bottle manipulator.}
\noindent Palm: $R_{p}1, R_{p}6, R_{p}6, R_{p}5, R_{p}7, R_{p}4, R_{p}5, R_{p}5, R_{p}7, R_{p}7, \allowbreak R_{p}7$.\\
Upper (abduction/adduction) finger: $R_{f}3, R_{f}5, R_{f}14, R_{f}14, \allowbreak R_{f}14, R_{f}15, R_{f}2.$ \\
Lower (flexion/extension) fingers: $R_{f}3, R_{f}7, R_{f}5, R_{f}14, \allowbreak R_{f}14, R_{f}14, R_{f}15, R_{f}2.$\\
After developing all fingers: $R_{f}2$.

\subsection{Scissor manipulator.}
\noindent Palm: $R_{p}1, R_{p}6, R_{p}5, R_{p}6, R_{p}7, R_{p}6, R_{p}7$.\\
For each finger: $R_{f}3, R_{f}7, R_{f}8, R_{f}12.$\\
After developing all fingers: $R_{f}2$.

\section{Cage Deformation: High and Low Resolution Meshes}
\label{app:meshes}

\begin{figure}
    \centering
    \includegraphics[width = \columnwidth]{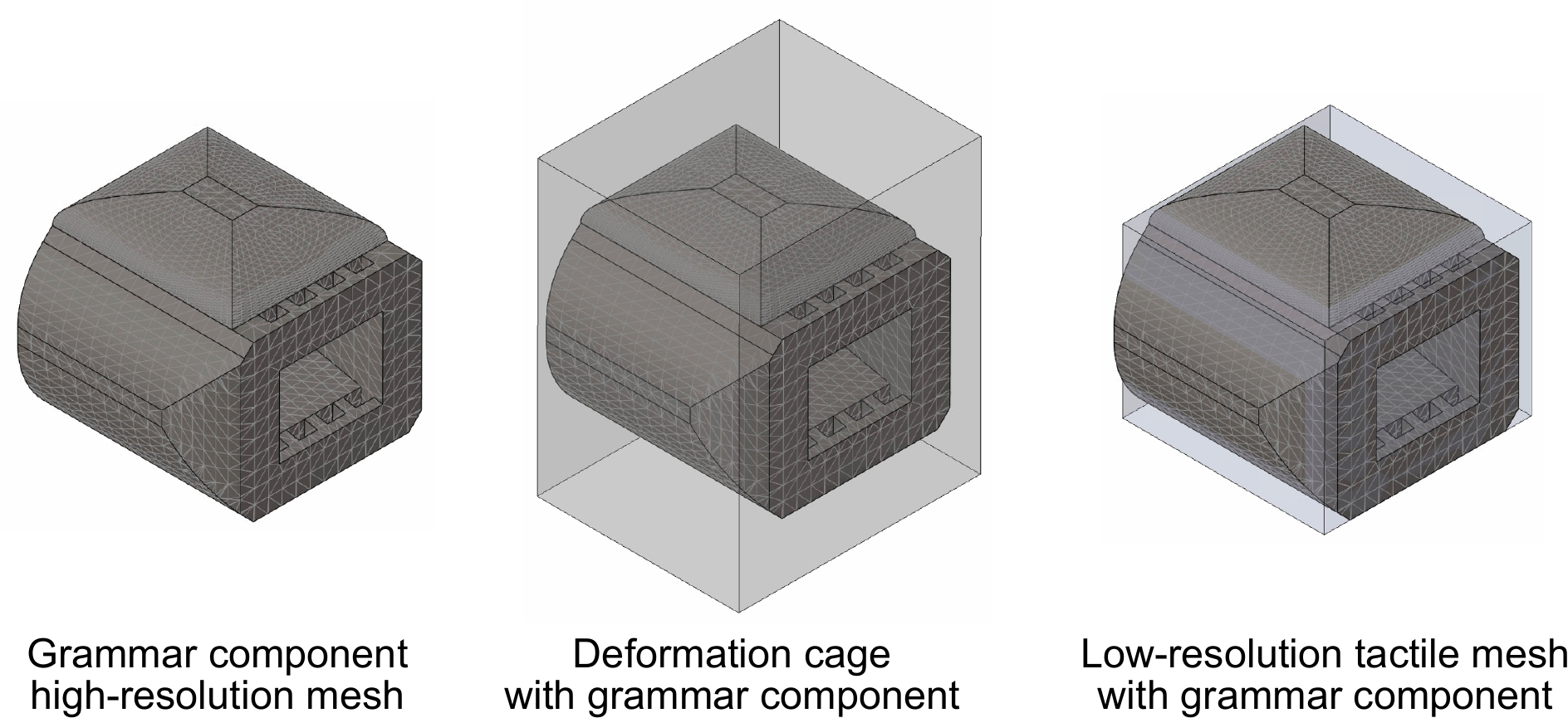}
    \caption{Each component has three meshes: a high-resolution mesh of the component used for 3D printing (left), a cuboid, low-resolution "cage" mesh (middle) to specify geometric deformation, and a coarse, low-resolution mesh (right) to generate knitting patterns. Here, the phalanx component is used as an example to compare these meshes. The grammar component high resolution mesh is shown in all three images for scale.} 
    \label{fig:mesh_comparison}
\end{figure}

Multiple high and low resolution meshes are involved in the design pipeline. Specifically, there are three meshes (shown in Fig. \ref{fig:mesh_comparison}) associated with each grammar component:
\begin{itemize}
    \item \textbf{A cuboid, low-resolution "cage" mesh} fully enclosing each grammar component. The vertices of this cage are used to define the basis of the deformation, and the user moves the cage vertices to deform the robotic manipulator components. These points are the only points the user has control over when altering the geometry of the manipulator: all other meshes deform according to the user-specified cage mesh deformations.
    \item \textbf{A high-resolution mesh} of each grammar component. Once combined to form a manipulator, these meshes will be 3D printed. This mesh is affected by deformations the user applies to the cage: when the user widens one end of the cage, the corresponding end of the high-resolution grammar component mesh will also widen.
    \item \textbf{A coarse, low-resolution mesh} used to generate knitting patterns. This is a highly simplified version of the grammar component and is sized so that the knitted pattern generated from this mesh fits snugly over the corresponding 3D printed part. These meshes are also subjected to the deformations the user applies to the deformation cage.
\end{itemize}

\begin{figure}
    \centering
    \includegraphics[width = \columnwidth]{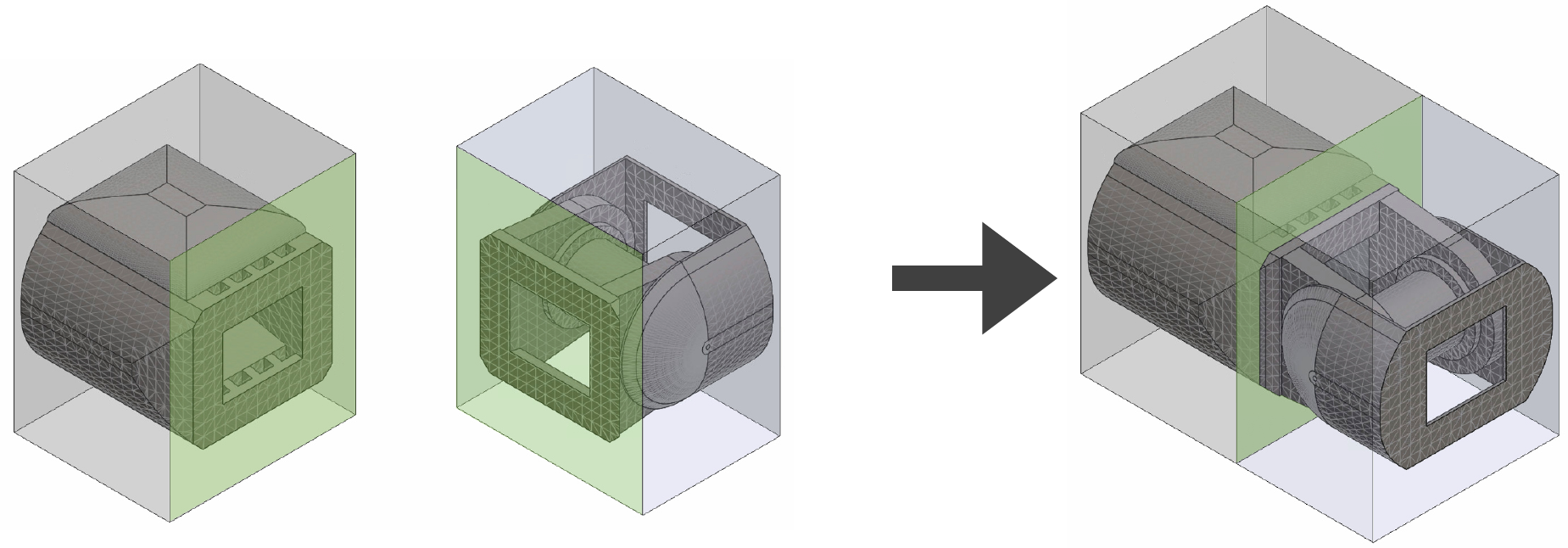}
    \caption{The component deformation cages are sized so that their vertices are exactly aligned when the components are joined.} 
    \label{fig:mesh_joining}
\end{figure}

Special considerations beyond those listed above had to be taken in determining the dimensions of the low resolution cages for each component. Let the face of a grammar component that connects to another grammar component be called the \textit{mating face}. Then, for every two grammar components that share a mating face, the vertices of the mating faces corresponding to those components must match for their two cage deformation meshes. This concept is illustrated in Fig. \ref{fig:mesh_joining}. Because the cage meshes must fully enclose all components and because cuboid mushes must always have aligning vertices, the height of the mesh must correspond to the largest component in the system. Similarly, the vertices of the mating faces of the tactile meshes of two components must also match. However, these tactile meshes are not required to fully enclose the high resolution mesh, nor are they required to be perfectly cuboid.

Why must this vertex-matching constraint exist? For the cage deformation mesh, the vertices of two mating faces must match because, upon mating, the overlapped vertices merge into one point that controls the two components on either side of the point. This ensures that the interface between the two components and any feature that spans those pieces remains intact. This principle guarantees both a watertight mesh and manufacturability after deformations. For the knitting mesh, the vertices of the two mating faces must align to form one continuous knitting surface. If the two faces did not match perfectly, there would be a sort of "step" between the smaller and bigger faces, causing the knitting program to generate a cover with a step and corners that does not reflect any feature of the high-resolution mesh. To maintain smooth continuity, the two mating faces must perfectly align.


\section{Control: Finite State Machine}
\label{app:fsm}

\begin{figure}[ht]
    \centering
    \includegraphics[width = \columnwidth]{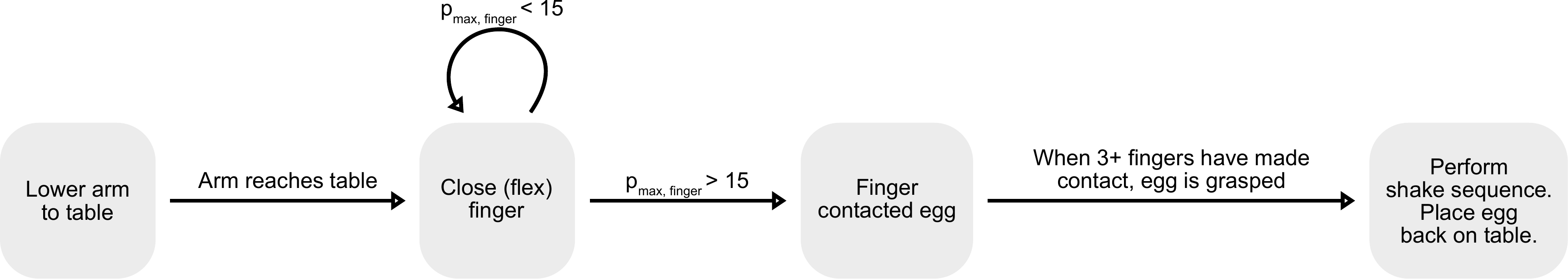}
    \caption{Finite state machine diagram for egg picker control sequence.} 
    \label{fig:fsm_egg}
\end{figure}

\begin{figure}[ht]
    \centering
    \includegraphics[width = \columnwidth]{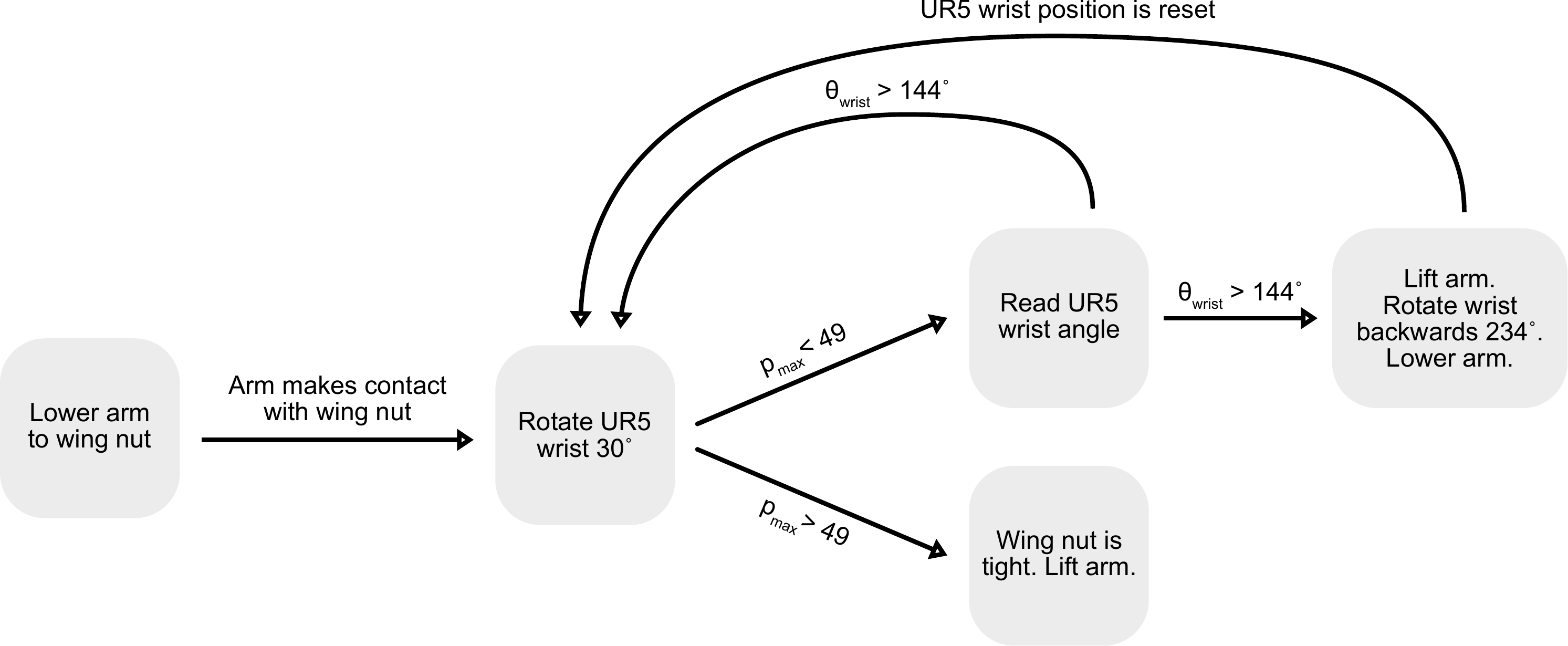}
    \caption{Finite state machine diagram for wing screw tightening control sequence. In the diagram, $\theta_{wrist}$ refers to the angle of the most distal joint on the UR5 arm.} 
    \label{fig:fsm_wing}
\end{figure}

\begin{figure}[ht]
    \centering
    \includegraphics[width = \columnwidth]{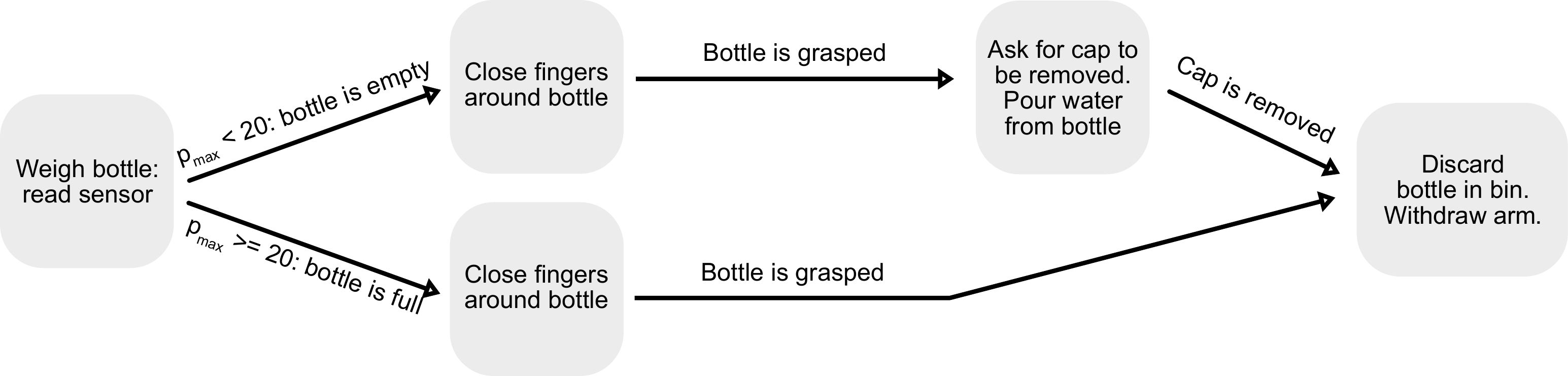}
    \caption{Finite state machine diagram for water bottle sorting control sequence.} 
    \label{fig:fsm_bottle}
\end{figure}

\begin{figure}[ht]
    \centering
    \includegraphics[width = \columnwidth]{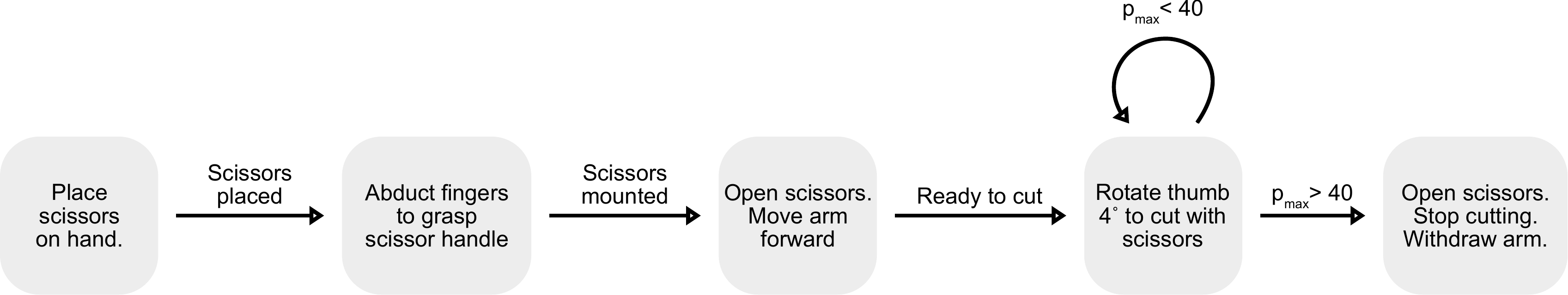}
    \caption{Finite state machine diagram for scissor cutting control sequence.} 
    \label{fig:fsm_scissor}
\end{figure}

Finite state machine controls based on sensor inputs were used to perform the manipulator tasks. These controls are not novel and were only intended to demonstrate that the manipulators are easily controllable; therefore, the positions of objects in the tasks were hard coded rather than determined intelligently via, for example, computer vision. In each of the state machine diagrams (Figs. \ref{fig:fsm_egg} - \ref{fig:fsm_scissor}), $p_{max}$ refers to the maximum reading in a sensor patch on a manipulator finger.

\section{Manipulator Performance}
\label{app:sensor_graphs}

\begin{figure}[ht]
    \centering
    \includegraphics[width = \columnwidth]{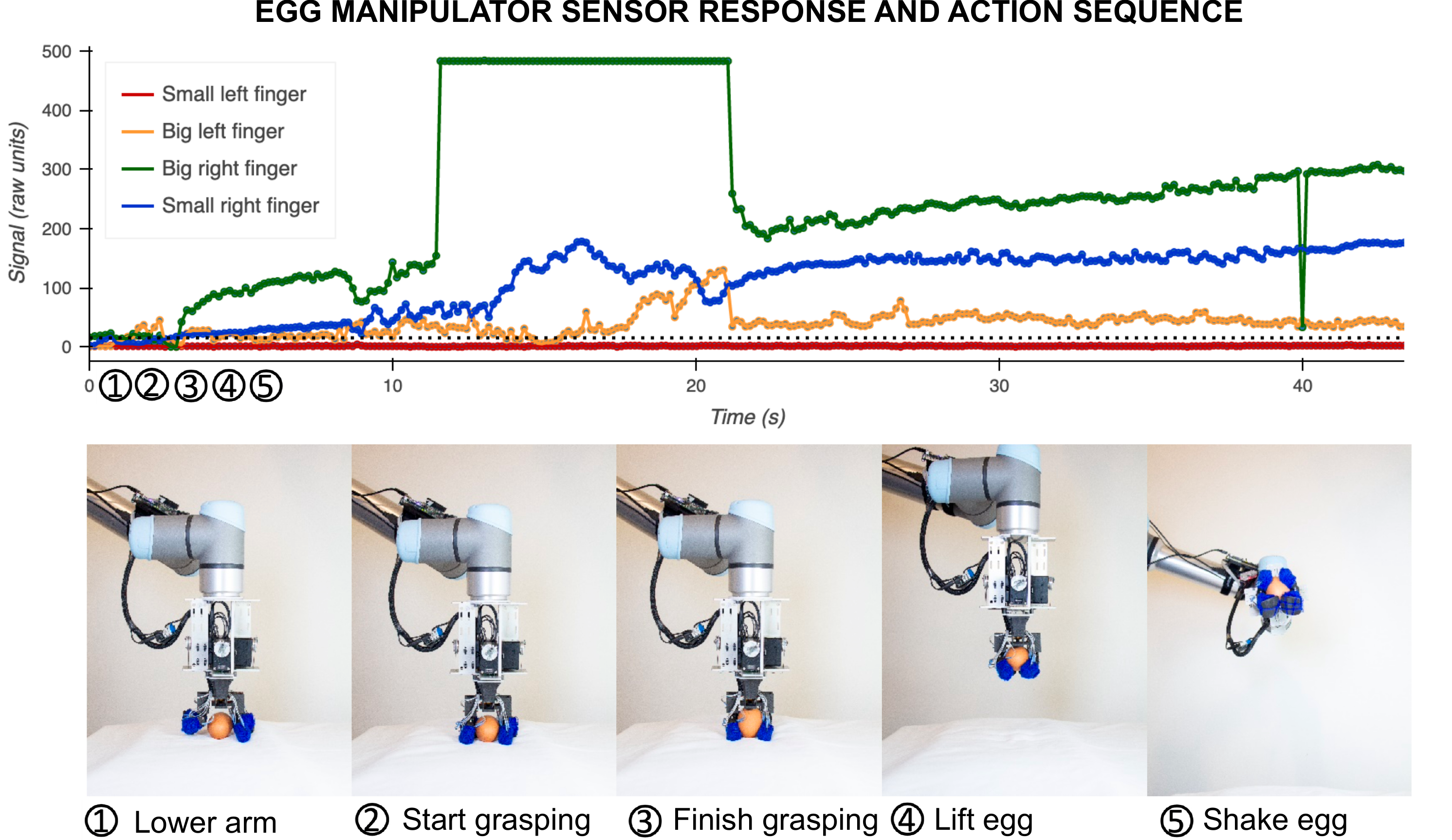}
    \caption{A typical egg picking action sequence with the maximum readings from the tactile sensors on each finger. The grasp registers as successful if three out of the four finger readings exceed our set threshold.}
    \label{fig:app_eggpicker_tactile}
\end{figure}

\begin{figure}[ht]
    \centering
    \includegraphics[width = \columnwidth]{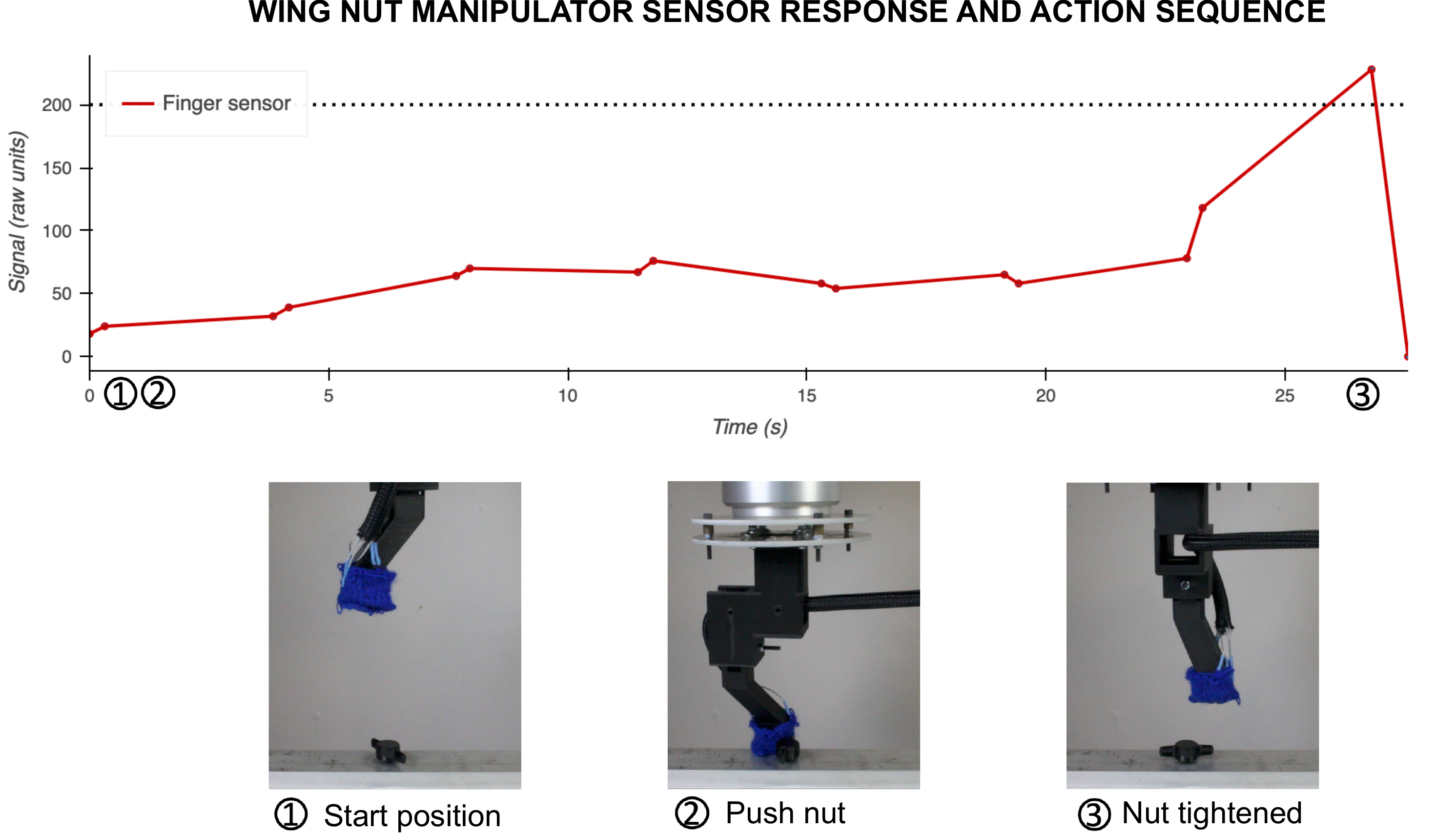}
    \caption{A typical wing screw tightening action sequence of the manipulator with the maximum readings from the tactile sensors on the finger. The wing screw is registered as tightened if it exceeds our set threshold.} 
    \label{fig:app_wing_tactile}
\end{figure}

\begin{figure}[ht]
    \centering
    \includegraphics[width = \columnwidth]{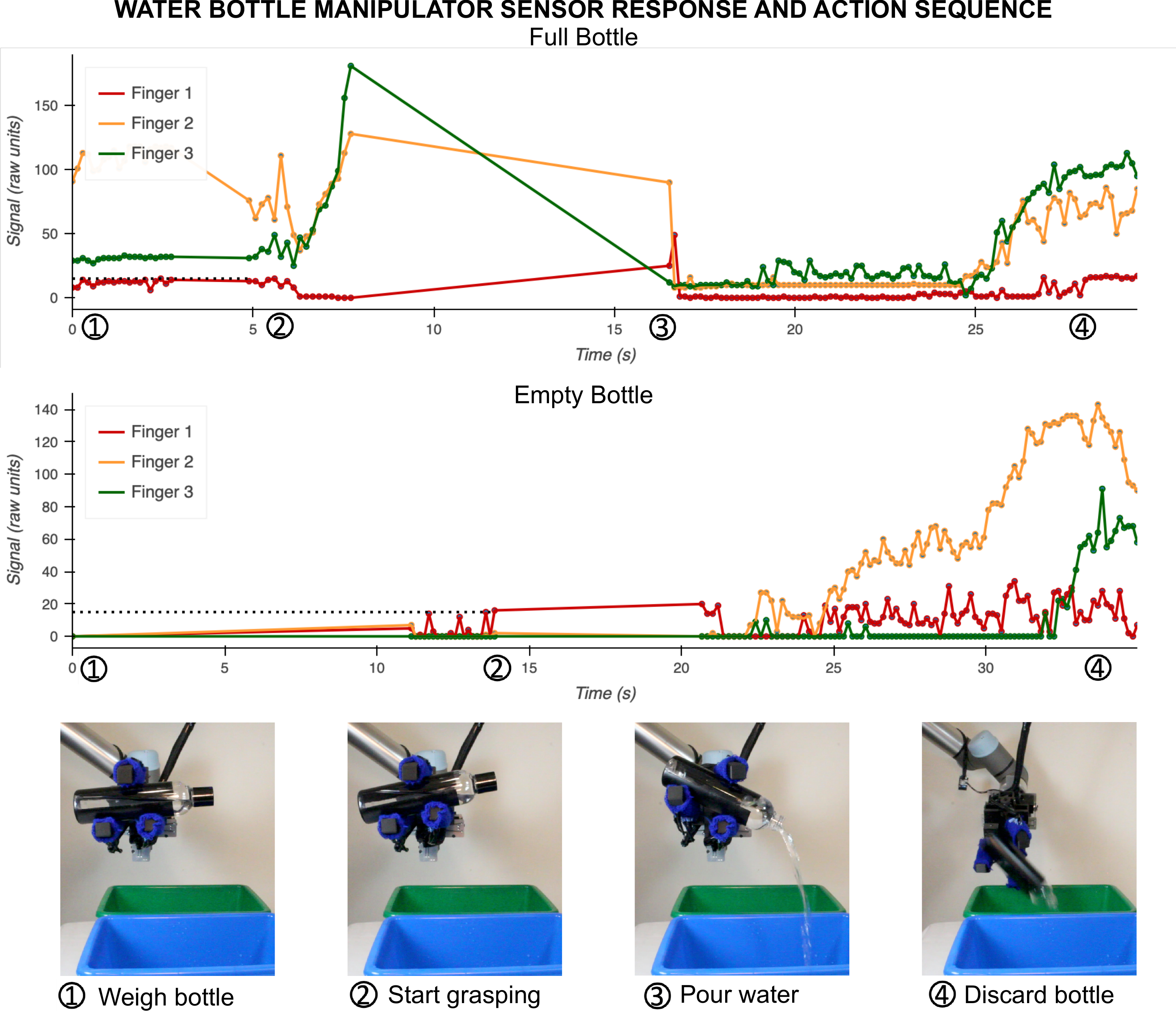}
    \caption{A typical bottle task action sequence of the manipulator with the maximum readings from the tactile sensors on each finger. The water bottle registers as full if the pressure on either lower finger exceeds the threshold.} 
    \label{fig:app_bottle_tactile}
\end{figure}

\begin{figure}[ht]
    \centering
    \includegraphics[width = \columnwidth]{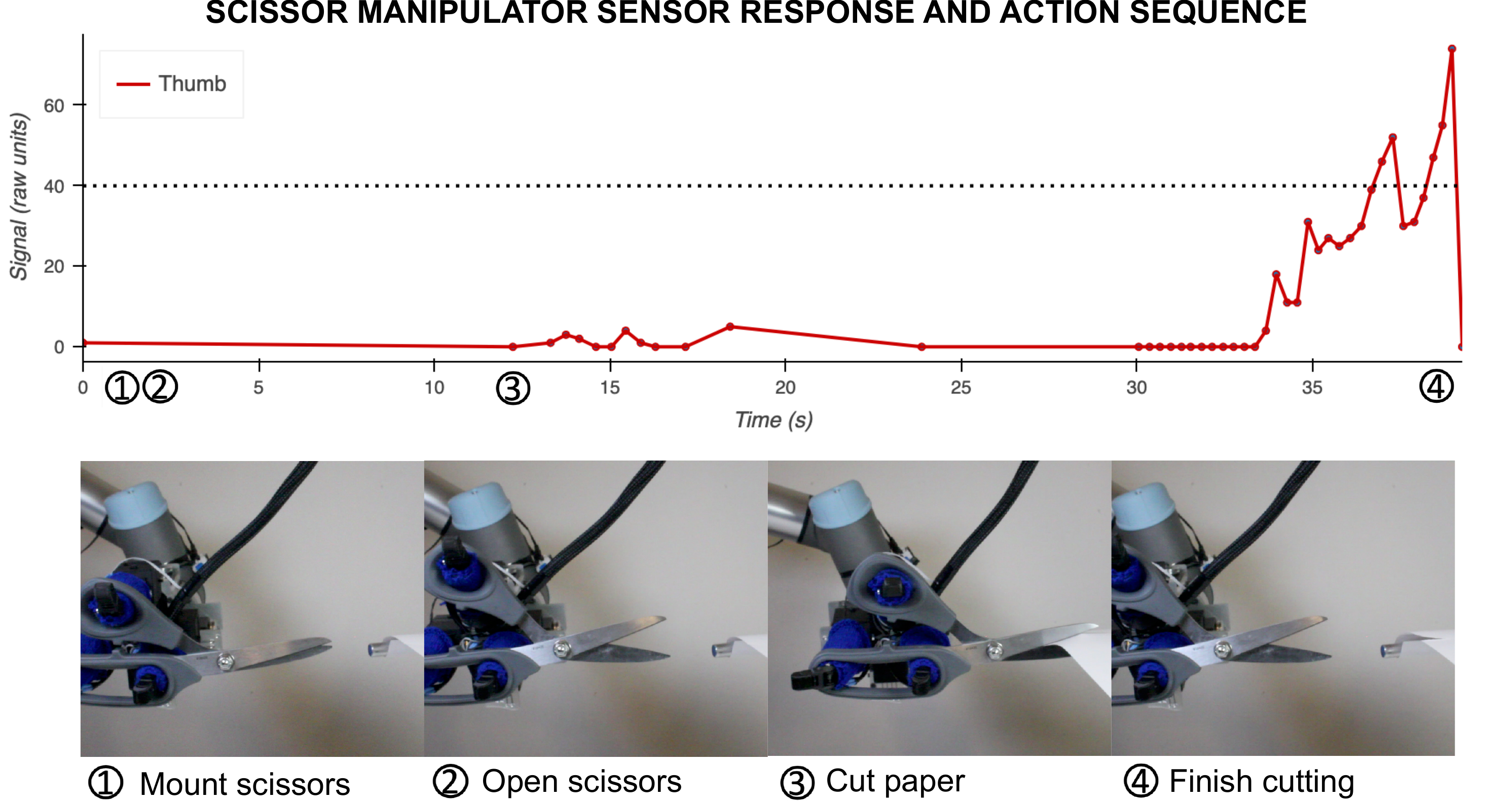}
    \caption{A typical cutting action sequence of the manipulator with the maximum readings from the tactile sensors on the manipulator's thumb. The cut registers as "completed" if the pressure exceeds the threshold even if the scissors are not closed; this protects the motors from overload.} 
    \label{fig:app_scissor_tactile}
\end{figure}

This section contains image sequences with sensor data for each manipulator and task (Figs. \ref{fig:app_eggpicker_tactile} - \ref{fig:app_scissor_tactile}). For each task, the threshold required to perform the task was experimentally determined. For instance, for grasping tasks, the threshold was chosen to correspond to the strength of grasp such that the object does not slip out of the hand. Please note that the sensor data plots are a depiction of a single run and are intended as only an approximate illustration of sensor responses while executing tasks. 


\end{appendices}
\end{document}